# An Improved Gauss-Newtons Method based Back-propagation Algorithm for Fast Convergence

Sudarshan Nandy
DETS, Kalyani University,
Kalyani, Nadia, West Bengal.

Partha Pratim Sarkar
DETS, Kalyani university,
Kalyani, Nadia, West Bengal.

Achintya Das
Kalyani Govt. Engineering
College,Kalyani, Nadia, West Bengal.

## ABSTRACT
The present work deals with an improved back-propagation algorithm based on Gauss-Newton numerical optimization method for fast convergence. The steepest descent method is used for the back-propagation. The algorithm is tested using various datasets and compared with the steepest descent back-propagation algorithm. In the system, optimization is carried out using multilayer neural network. The efficacy of the proposed method is observed during the training period as it converges quickly for the dataset used in test. The requirement of memory for computing the steps of algorithm is also analyzed.

## 1. INTRODUCTION
In neural network, training the learning algorithms plays quite important role in the process. Throughout the process, the learning algorithm is used to adjust the weight, bias and other input parameters in such a manner that the model is able to count its best fit with the environment in a minimum amount of time. Back-propagation in multilayer neural network is one of the supervised training procedure and variety of approaches are developed over the last decade for its applicability to a numerous number of application[12][13].

The steepest descent back-propagation (SDBP) is used in several applications despite its asymptotic slow convergence rate [1][3]. The algorithm is also known as a gradient method. The slow convergence rate of steepest descent algorithm encourages many ides to be developed for faster convergence rate in training a multilayer neural network. The methods that are developed may be divided into two categories. In first category of work the parameters are updated heuristically[2][4][5][6]. The momentum based back-propagation is one of the methods under this category where the momentum factor is fixed, being it always less than one. Another type of back-propagation under this category is variable learning back-propagation where the momentum and learning rates are adjusted in every iteration. In this category the parameters are updated in an ad-hoc manner. In some cases the heuristically modification to the parameters is required to converge the back-propagation algorithm quickly, but the algorithms under this category update several parameters, while only learning rate is related to the steepest descent back-propagation. Another problem of this type of approach is that the values of those parameters are problem oriented.

Numerical procedures are used in another categories to optimize the steepest descent back-propagation. The method uses various optimization techniques among which Levenberg-Marquardt back-propagation algorithm and conjugate back-propagation algorithm are used in various applications. The approach to second category is quite popular and widely applied in various applications[7]. The conjugate back-propagation method is also known as quasi-newton method. Besides efficiency of quasi-newton methods, the algorithm suffers from storage and computational requirements, increasing more with complexity of neural network design.

The other most popular numerical optimization method is Levenberg-Marquardt algorithm [8][9][10][11][15]. The algorithms for least square estimation of non-linear parameters work on the basis of two approaches. First approach is Taylor series approach where modification to all parameters are performed on each iteration. Second approach is a gradient method or steepest descent method. The Levenberg-Marquardt algorithm provides an interpolation between the two approaches, and the algorithm depends on the initial non-linear input parameters. This technique is used to calculate the damping factor, and the whole algorithm basically depends on the value of initial damping , damping increment, damping decrement and minimum damping. The initial values of those factors are responsible to converge the algorithm more quickly or slowly. The stability and rate of convergence of this algorithm is affected by the damping parameters. This algorithmic process is a problem for a system where the behavior of the environment changes dynamically. The algorithm is also not a memory efficient in case of training a large multilayer neural network.

The algorithms for least square estimation of non-linear parameters usually update the parameters in sequential and batch mode. In case of sequential parameters, updation of weight , bias and other parameters is modified on each representation of input/output pair. In batch mode operation, input parameters are updated on each operation.

The proposed algorithm is a developed method based on the Gauss-Newton numerical optimization technique. It resulting a quick convergence with stability in case of multilayer neural network learning. The proposed algorithm is also analyzed for its memory requirement during the time of training on datasets. Convergence criteria is another important issue for neural network training algorithm. Convergence criteria used not only terminate the algorithm but also help to compare with other algorithm. The proposed method converges according to the following criteria:

1. MSE (Mean Square Error): The algorithm stops when it reaches the pre-specified threshold value.

2. Correct Classification: The main idea behind the convergence criteria is that if any neural network training algorithm correctly classifies some input, then it is possible to calculate the percentage of correctly classified input, and the same is increased or decreased according to the ability of training algorithm. The proposed algorithm is designed with one pre-defined correct classification threshold value.

## 2. PROPOSED METHOD

The algorithm to train a feed forward network with back-propagation method is required to discuss the proposed method . The discussion and analysis on normal back-propagation method and proposed method are studied with the 3-3-1 designed feed-forward neural network(Fig. 1.).

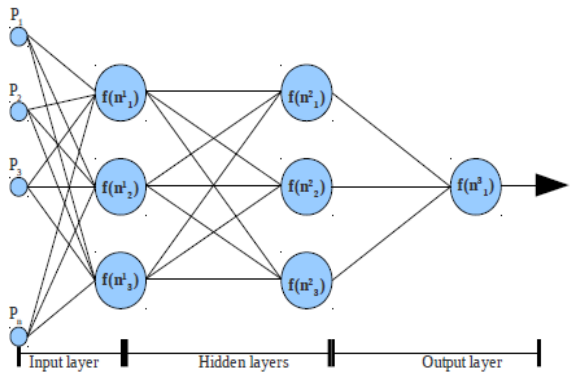

**Fig.1. 3-3-1 design of feed-forward Neural network.**

In case of multilayer networks as depicted in the Fig.1 the net input of an unit u of layer n+1 is

$$N^{n+1}(u) = \sum_{j=1}^{n} W^{(n+1)}(u,j) a^n(u) + B^{(n+1)}(u)$$

.........(1)

where j = 1, 2,3,......., m

N= net input

W = Input weight

B= bias of a neuron

a = output from a neuron

So the final output form the layer is calculated as:

$$a^{(n+1)}(u) = f^{(n+1)}(N^{(n+1)}(u)) \quad ......(2)$$

The transfer function is denoted as $f^{(n)}$.

Now the algorithm is started with previously specified pattern considered as:

$$\{(p_1,t_1),(p_2,t_2),(p_3,t_3)............(p_n, t_n)\}$$

'p' being the input pattern and 't' the target associated with each pattern. The performance index considered here is the mean of squared error or M(x). So the performance index is calculated using following formula:

$$M(x) = \sum_{(j=1)}^{(m)} (t_j - a_j)^T (t_j - a_j) \quad (3)$$

where 'x' is represented as a vector of network weight and bias. Now if it is considered that $q(x) = (t_j - a_j)$ then Equ.(3) can be written as:

$$M(x) = \sum_{(j=1)}^{(m)} q(x)^T . q(x) \quad (4)$$

So, the weight and bias are calculated as follows

$$w_{(u,j)}^{(n+1)} = w_{(u,j)}^{(n)} - \alpha . \frac{d\hat{M}}{dw_{(u,j)}^{(n)}} \quad (5)$$

and

$$B_{(u,j)}^{(n+1)} = B_{(u,j)}^{(n)} - \alpha . \frac{d\hat{M}}{dB_{(u,j)}^{(n)}} \quad (6)$$

Now, if $N_{(u,j)}^n = \sum_{(j=1)}^{(n-1)} W_{(u,j)}^n . a_j^{(n-1)} + B_{(u,j)}^n$

then

$$\frac{dN_u^n}{dW_{(u,j)}^n} = a^{(n-1)} \quad \text{and} \quad \frac{dN_u^n}{dB_{(u,j)}^n} = 1$$

If it is considered that

$$\frac{d\hat{M}}{dN_u^n} = s^{(n)}$$

then Equ.(5) and (6) can be written as:

$$W_{(u,j)}^{(n+1)} = W_{(u,j)}^n - \alpha.s^n (a^{(n-1)})^T \quad (7)$$

and

$$B_{(u,j)}^{(n+1)} = B_{(u,j)}^n - \alpha.s^n \quad (8)$$

Here α is a stable learning rate. $S^n$ is the sensitivity at layer 'n' calculated from sensitivity at the layer (n+1). The sensitivity of the layer is calculated in backward i.e. the calculation of sensitivity is started from the last layer and ends at first layer of network. The sensitivity of the network is calculated as follows :

$$S^n = f^n(N^n)(W^{(n+1)})^T . s^{(n+1)} \quad (9)$$

The sensitivity of the starting layer is calculated as

$$S^n = -2.f^n(N^n)(t_j - a_j) \quad (10)$$

where,

$$f^n(N^n) = \begin{bmatrix} f^n(N_1^n) & 0 & 0 \\ 0 & f^n(N_2^n) & 0 \\ \cdots \cdots \cdots \\ 0 & 0 & f^n(N_k^n) \end{bmatrix}$$

and

$$f^n(N) = \frac{d(f^n(N))}{dN}$$

The gradient is the first-order derivative of the mean of square error. The gradient is calculated as follows

$$\nabla M(x) = \left[ \frac{d(M(x))}{dx_1}, \frac{d(M(x))}{dx_2}, \cdots \frac{d(M(x))}{dx} \right]$$

from Equ.(4) the gradient is calculated

$$\nabla M(x) = 2 \sum_{(j=1)}^{(m)} q_j(x) \cdot \frac{dq_j(x)}{dx}$$

as

$$\nabla M(x) = 2J^T(x).q(x) \quad (11)$$

where J(x) is the Jacobian matrix and it is written as

$$J(x) = \begin{bmatrix} \frac{dq_1(x)}{dx_1}, \frac{dq_1(x)}{dx_2} \cdots \cdots \frac{dq_1(x)}{dx_n} \\ \frac{dq_2(x)}{dx_1}, \frac{dq_2(x)}{dx_2} \cdots \cdots \frac{dq_2(x)}{dx_n} \\ . \quad , \quad . \quad \cdots \cdots \quad . \\ . \quad , \quad . \quad \cdots \cdots \quad . \\ . \quad , \quad . \quad \cdots \cdots \quad . \\ \frac{dq_k(x)}{dx_1}, \frac{dq_k(x)}{dx_2} \cdots \cdots \frac{dq_k(x)}{dx_n} \end{bmatrix}$$

On each representation of input parameters the network generates the $S^M$ number of error. So,

$$S^L = [s_1^L, S_2^L, \ldots, S_n^L]$$

One error in $S^L$ matrix is used to creates one row in the Jacobian matrix. So, two element in row of Jacobian matrix is created for one representation of input parameters to the network.

The Hessian matrix is the double derivative of the performance index. It is calculated as follows

$$\nabla^2 M(x) = \begin{bmatrix} \frac{d^2(M(x))}{dx_1^2}, \cdots \cdots, \frac{d^2(M(x))}{dx_1 dx_n} \\ \frac{d^2(M(x))}{dx_1 dx_2}, \frac{d^2(M(x))}{dx_2^2}, \cdots \cdots \\ . \quad . \quad , \cdots \cdots \\ . \quad . \quad , \cdots \cdots \\ . \quad . \quad , \cdots \cdots \\ \frac{d^2(M(x))}{dx_n dx_1}, \cdots \cdots, \frac{d^2(M(x))}{dx_n} \end{bmatrix}$$

The elements of Hessian matrix r, l are calculated as

$$\nabla^2 M(x) = 2 \sum_{(j=1)}^{(m)} \frac{dq_j(x)}{dx_r} \cdot \frac{dq_j(x)}{dx_l} + q_j(x) \cdot \frac{d^2 q_j(x)}{dx_r dx_l}$$

$$\nabla^2 M(x) = 2J^T(x).J(x) + 2v(x) \quad (12)$$

if v(x) is too small then

$$\nabla^2 M(x) = 2J^T(x)J(x) \quad (13)$$

Now according to the Newtons formula

$$\Delta x = -H^{-1}.G_k \quad (14)$$

where H is the Hessian matrix and $G_k$ is the gradient. So if Equ.(13) and (11) substitute into newtons formula (Equ. (14)) then the formula is

$$\Delta X = -[2J^T(x)J(x)]^{-1}.[2J^T(x).q(x)]$$

$$\Delta X = -[J^T(x)J(x)]^{-1}.J^T(x).q(x) \quad (15)$$

The Equ.(15) is the Gauss-Newton method for optimization of performance index.

The proposed algorithm includes one adjustment to the weight and bias after every iteration. The adjustment is computed for each weight and bias of each layer. The adjustment parameter is defined as:

$$J^T(x)q(x) = \frac{1}{2}\nabla M(x)$$

The following is the formula to adjust weight and bias

$$W_{(u,j)}^{(n+1)} = W_{(u,j)}^{n} - \frac{1}{2}\nabla M(x) \quad (16)$$

and

$$B_{(u,j)}^{(n+1)} = B_{(u,j)}^{n} - \frac{1}{2}\nabla M(x) \quad (17)$$

The proposed algorithm for fast convergence of back-propagation algorithm thus goes as follows:

Begin:

[1] Calculate M(x) after all inputs are represented to the neural network using Equ.(3).

[2] Calculate the Jacobian matrix J(x) and gradient $G_k$.

[3] Now calculate the ΔX using Equ.(15) and adjust the weight and bias using Equ.(16) and (17).

[4] Recompute the performance index and compare with the performance index calculated in step 1. If the newly calculated performance index is less than the step 1, then check the convergence criteria, and if the convergence is less then the pre-defined threshold value, then stop the algorithm, otherwise goto step 1.

The proposed algorithm introduces a new step where the weight is adjusted before each iteration. In some cases it is observed that the proposed algorithm results fast convergence. The requirement of the memory for the whole process of optimization is also minimum. The algorithm is found to behave like normal optimization, if weight adjustment is not performed. It is possible to convert the same algorithm into Levenberg-Marquardt algorithm, if one damping factor is introduced in the calculation of ΔX, and is adjusted according to the requirement of algorithmic optimization criteria.

## 3. EVALUATION

The performance of proposed algorithm is evaluated in terms of percentage of correct classification, mean of square error and the number of iteration required to converge. The algorithm is also analyzed for its memory requirement and it is compared with the back-propagation algorithm based on steepest descent method.

### 3.1 Experimental set-up

In order to test the algorithm, a system is designed and coded in the python programming language. The experiment is performed on a Pentium-IV core 2 Duo 1.66 GHz. processor based machine configured with the 512 MB of RAM space.

The data set are classified with its own classifier built inside of the main program. The dataset used to test the proposed method are

1. Iris data Set

2. Wine Data Set.

The iris data set [16] consists of 150 number of instances, and each instance consists of 4 numbers of attribute. Attributes are mainly the features of the iris flower and the data set consists of following features of the iris flower

a. Sepal length

b. Sepal width

c. Petal length

d. Petal width.

On the basis of those features the data set consists of three classes and they are 1. Iris Setosa 2. Iris Versicolour 3. Iris virginica. Those are the various types of iris flower. The data in the Iris data set are non-linear.

The Wine data set [17] consists of 178 instances of data and each data instance consists of 13 number of attributes. Those attributes are the different chemical compound of three types of wine. The chemical composition of the wine is the feature of the wine.

The system control flow is depicted in Fig.2. The non-linear input parameters are processed through the classifier and data processing unit is required to finalize the data before training and testing. The data processing unit prepared the various parameter is needed to train the neural network, and is connected to the 'anncfg.conf' file. The unit is responsible to design the network according to the setting defined in the file. The type of algorithm is used for analysis defined in the file. The next part is the "ANN BOX" which is the main unit of the system. The proposed method is implemented in this unit.

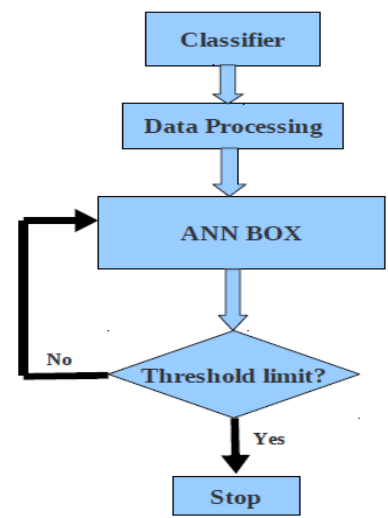

**Fig. 2. The system control-flow diagram.**

The system stops when any of the previously defined convergence criteria crosses the threshold limit.

## 3.2 Experimental result

The experiment with the new proposed method is performed on the basis of mean square error, percentage of correct classification, memory requirement and number of iterations required to stabilize the calculation. The summary of all observed data and graphs found at the time of experiment is depicted here.

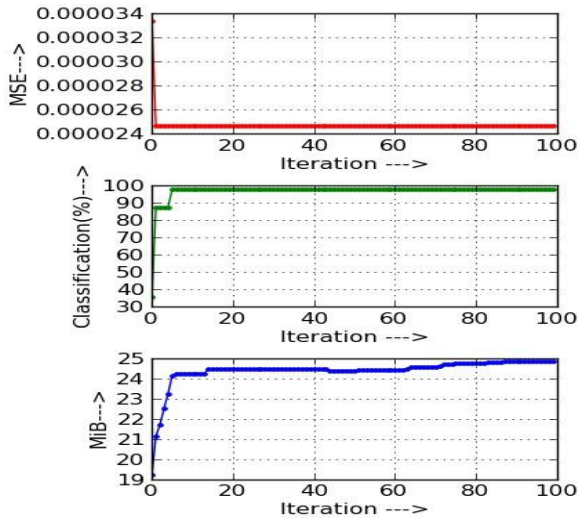

**Fig. 3. Iris data set training using Steepest Descent back-propagation.**

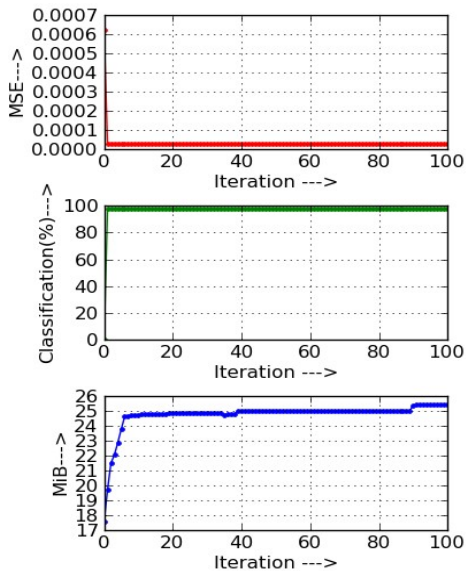

**Fig. 4. Iris data set training using proposed method.**

**Table 1. Observed convergence parameter for Iris data set training.**

| Sl. No | Convergence parameters | Steepest Descent Back-propagation (SDBP) | Improved Gauss-Newton Method based algorithm |
|---|---|---|---|
| 1 | Mean of Squared Error (MSE) | 2.47E-005 | 2.47E-005 |
| 2 | Correct Classification(%) | 97.78% | 97.78% |
| 3 | Memory required(MiB) | 25 | 25.5 |
| 4 | No. iteration required to stable at high Classification | 5 | 2 |

The experimental result reflects the fast convergence nature of the Gauss-Newton based proposed method. The requirement of the memory is also low in comparison with the proposed method if the convergence criteria is based on the correct percentage of classification or MSE rate. The steepest descent method requires five iterations to stabilize the calculation and the required memory at that iteration is 23 to 23.5 MiB. The proposed method is stable on its second iteration and hence the required memory is 19.9 to 20.5 MiB.

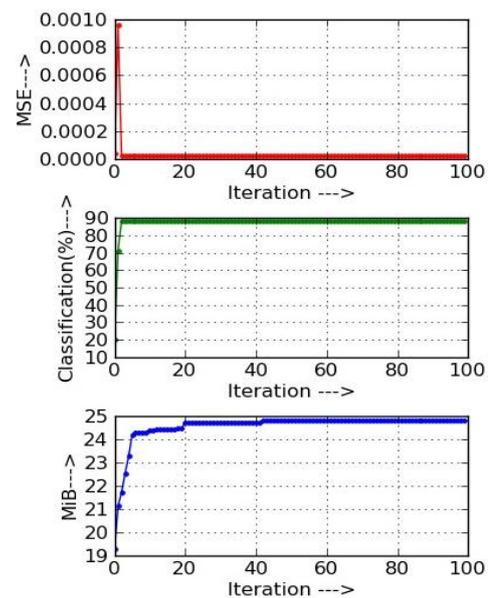

**Fig. 5. Wine data set training based on Steepest Descent back-propagation method.**

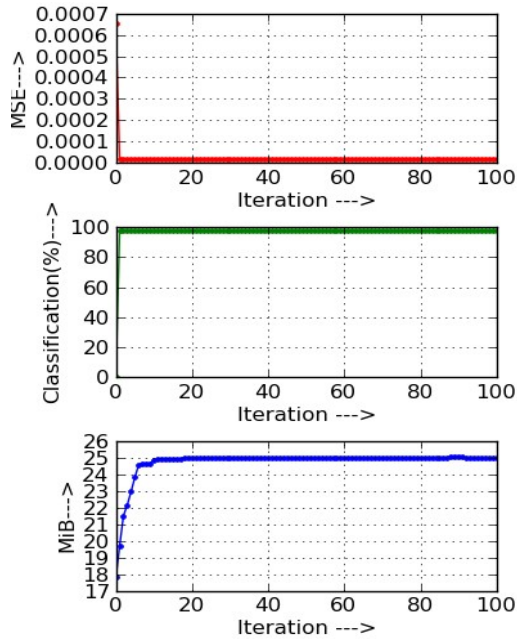

**Fig. 6. Wine data set training with on improved Gauss-Newton based method.**

**Table 2. Observed convergence parameter for Wine data set training.**

| Sl. No | Convergence parameters | Steepest Descent Back-propagation (SDBP) | Improved Gauss-Newton Method based algorithm |
|---|---|---|---|
| 1 | Mean of Squared Error (MSE) | 1.824E-005 | 1.824E-005 |
| 2 | Correct Classification(%) | 88.54% | 98.089% |
| 3 | Memory required(MiB) | 24.9 | 25 |
| 4 | No. iteration required to stable at high Classification | 100 | 2 |

This experiment shows that the wine data set is much more correctly classified with the proposed method. The proposed algorithm converges and became stable after second iteration, whereas the steepest descent method may take long time. The memory required for steepest descent is 24.9MiB but the proposed method requires only 18 to 19.9MiB of memory if its convergence is strictly based on correct classification or mean of squared error (MSE).

## 4. CONCLUSION

An improved method based on the Gauss-Newton numerical optimization technique is described and the experimental study with two no-liner data set are presented. The experimental result discussed in the above section based on the proposed method and one conventional method, it may be conclude that the proposed technique is superior in terms of time taken to converge and the memory space, both being reduced to a great extent. Thus the benefit of the present method is established.